\documentclass[conference]{IEEEtran}
\IEEEoverridecommandlockouts
\usepackage{afterpage}

\usepackage{algorithmic}
\usepackage{algorithm}
\usepackage{balance}

\usepackage{cite}
\usepackage{amsmath,amssymb,amsfonts}

\usepackage{arydshln}  
\usepackage{graphicx}
\usepackage{multirow}
\usepackage{subcaption} 
\usepackage{textcomp}
\usepackage{tikz,hyperref}
\usepackage{booktabs}
\usepackage{array}
\usepackage[table]{xcolor}
\newcolumntype{L}[1]{>{\centering\arraybackslash}m{#1}}
\newcolumntype{C}[1]{>{\centering\arraybackslash}p{#1}}

\usepackage{xcolor}
\AtBeginDocument{%
  \setlength{\abovedisplayskip}{4pt}%
  \setlength{\belowdisplayskip}{4pt}%
  \setlength{\abovedisplayshortskip}{3pt}%
  \setlength{\belowdisplayshortskip}{4pt}%
}
\def\BibTeX{{\rm B\kern-.05em{\sc i\kern-.025em b}\kern-.08em
    T\kern-.1667em\lower.7ex\hbox{E}\kern-.125emX}}

\definecolor{lime}{HTML}{A6CE39}
\DeclareRobustCommand{\orcidicon}{%
	\begin{tikzpicture}
	\draw[lime, fill=lime] (0,0) 
	circle [radius=0.16] 
	node[white] {{\fontfamily{qag}\selectfont \tiny ID}};
	\draw[white, fill=white] (-0.0625,0.095) 
	circle [radius=0.007];
	\end{tikzpicture}
	\hspace{-2mm}
}

\foreach \x in {A, ..., Z}{%
	\expandafter\xdef\csname orcid\x\endcsname{\noexpand\href{https://orcid.org/\csname orcidauthor\x\endcsname}{\noexpand\orcidicon}}
}

\begin{document}

\title{\LARGE \bf Alternating Bi-Objective Optimization for \\ Explainable Neuro-Fuzzy Systems}

\author{\IEEEauthorblockN{Qusai Khaled{\orcidA{}}}
\IEEEauthorblockA{\textit{Jheronimus Academy of Data Science} \\
\textit{Eindhoven University of Technology}\\
Eindhoven, The Netherlands \\
qusai.khaled@ieee.org}
\and
\IEEEauthorblockN{Uzay Kaymak{\orcidB{}}}
\IEEEauthorblockA{\textit{Jheronimus Academy of Data Science} \\
\textit{Eindhoven University of Technology}\\
Eindhoven, The Netherlands \\
u.kaymak@ieee.org}
\and
\IEEEauthorblockN{Laura Genga {\orcidC{}}}
\IEEEauthorblockA{\textit{School of Industrial Engineering} \\
\textit{Eindhoven University of Technology}\\
Eindhoven, The Netherlands \\
l.genga@tue.nl}}
\maketitle

\afterpage{%
  \addtolength{\topmargin}{-0.25in}%
  \addtolength{\textheight}{0.25in}%
}

\begin{abstract}
Fuzzy systems show strong potential in explainable AI due to their rule-based architecture and linguistic variables. Existing approaches navigate the accuracy-explainability trade-off either through evolutionary multi-objective optimization (MOO), which is computationally expensive, or gradient-based scalarization, which cannot recover non-convex Pareto regions. We propose X-ANFIS, an alternating bi-objective gradient-based optimization scheme for explainable adaptive neuro-fuzzy inference systems. Cauchy membership functions are used for stable training under semantically controlled initializations, and a differentiable explainability objective is introduced and decoupled from the performance objective through alternating gradient passes. Validated in approximately 5,000 experiments on nine UCI regression datasets, X-ANFIS consistently achieves target distinguishability while maintaining competitive predictive accuracy, recovering solutions beyond the convex hull of the MOO Pareto front.
\end{abstract}


\section{Introduction}

Recent developments in Explainable Artificial Intelligence (XAI) have renewed interest in fuzzy logic systems due to their inherent transparency through fuzzy rule-based architectures \cite{guillaume2001designing, gacto2010integration, moral2021explainable}. Fuzzy systems offer dual capabilities: linguistic interpretability via Mamdani-type systems and precise modeling through Takagi-Sugeno (T-S) frameworks \cite{takagi1985fuzzy}. When developed to be performance-oriented, fuzzy systems merge T-S systems with neural network learning capabilities without adhering to explainability considerations. T-S systems become similar to black box models in their opaque reasoning, all while achieving relatively high performance, which researchers refer to as precise fuzzy models \cite{jang1993anfis}. Linguistic fuzzy models, on the other hand, favor explainability over accuracy, prioritizing human-readable rule structures at the cost of predictive performance. The well-established accuracy-explainability trade-off \cite{arrieta2020explainable} has consequently motivated substantial research in interpretable fuzzy systems. With this mindset, explainability is maintained through preserving structural properties of the rule base and semantic properties of the fuzzy sets.

Existing methodologies navigate this landscape either after training through simplification techniques, such as pruning \cite{khaled2025interpretable}, merging \cite{setnes1998similarity}\cite{fuchs2020graph} and rule selection \cite{ishibuchi2002selecting}, or during training in multi-objective optimization (MOO) \cite{ishibuchi2004fuzzy}. The former essentially tunes a complex high-performing model with low explainability into a simpler lower-performance but more explainable model. The latter explores larger explainability-accuracy solution space by setting explainability as explicit objectives or constraints alongside accuracy, generating Pareto-optimal solutions that represent different accuracy-interpretability trade-offs. Among the different approaches to implement MOO, evolutionary algorithms, such as genetic algorithms and particle swarm optimization \cite{pulkkinen2009dynamically, marquez2010multi, shukla2012review}, are commonly combined with fuzzy systems when optimization is needed to balance explainability and accuracy. These global optimization techniques are better suited for fuzzy systems given the non-linearity of its membership functions. However, they are known to be computationally expensive.

Gradient-based methods such as weighted sum (scalarization) and $\epsilon$-constrained optimization \cite{de1999towards ,de2002semantic} are simpler more computationally efficient techniques, but they cannot identify Pareto-optimal solutions in the non-convex objective space of fuzzy systems, and their penalty factor requires careful tuning. Nevertheless, these methods can align coherently with the architecture of Adaptive Neuro-Fuzzy Systems (ANFIS), where gradients of differentiable membership functions (MF) guide optimization towards accurate models. While bearing potential and simplicity, limited research has focused on reshaping gradient-based optimization to address the explainability-accuracy trade-off in Neuro-fuzzy systems.

We propose an alternating bi-objective gradient-based optimization to achieve an explainable neuro-fuzzy system (X-ANFIS). In this architecture, both accuracy and interpretability objectives drive parameter optimization through decoupled gradient computations. The alternating gradient descent maintains a trajectory toward high accuracy while recursively optimizing explainability objectives. We evaluate ANFIS, X-ANFIS and weighted sum MO-ANFIS to investigate the potential of the proposed scheme. Nine real-world datasets from the UCI Machine Learning Repository are used for training, and the models are assessed using $R^2$, semantic interpretability, MSE, MAE and RMSE. While maintaining similar experimental conditions across the three models, a total of 5,032 models were trained, showing sensitivity to initializations, performances on the Pareto front and the distribution of membership functions.

The remainder of this paper is organized as follows. Section II briefly reviews T-S systems, fuzzy explainability taxonomies, related work and application challenges of gradient-based MOO in ANFIS. Section III presents the methodology of X-ANFIS. Section IV evaluates the proposed method on nine real-world datasets and two baseline models. Section V concludes the paper and outlines future research directions. The implementation code is publicly available at \url{https://github.com/QusaiKhaled/XANFIS}.

\section{Explainability of fuzzy systems}

Among the architectures of the most commonly used fuzzy systems, the T-S fuzzy system \cite{takagi1985fuzzy} approximates complex nonlinear relationships through a collection of local linear models, each activated by fuzzy membership functions. The system operates by partitioning the input space into fuzzy regions, each governed by a local linear model.

\subsection{Takagi-Sugeno System}
T-S fuzzy system with $m$ input variables and $R$ rules follows the structure:
\vspace{-0.1cm}
{\small
\begin{align}
R_r: & \text{IF } x_1 \text{ is } A_{1r} \text{ and } x_2 \text{ is } A_{2r} 
\text{ and } \ldots \text{ and } x_m \text{ is } A_{mr} \\
& \text{THEN } y_r = f_r(x_1, x_2, \ldots, x_m), \notag
\end{align}
}

\noindent
where $A_{ir}$ represents the $i$-th antecedent fuzzy set of rule $r$, and $f_r(\cdot)$ is the consequent function. Each $A_{ir}$ corresponds to a linguistic label such as \emph{low}, \emph{medium}, or \emph{high}, which can be defined based on expert knowledge or intuition about the variable $x_i$. This allows rules to be interpretable in human-understandable terms while still allowing precise computation.

\noindent
Assuming Gaussian membership functions for antecedents,

{\footnotesize
\begin{equation}
A_{ir}(x_i) = \exp\left(-\frac{(x_i - \mu_{ir})^2}{2\sigma_{ir}^2}\right),
\end{equation}
}

\noindent
where $\mu_{ir}$ and $\sigma_{ir}$ are the center and width parameters of the $i$-th antecedent in rule $r$. The firing strength of each rule is computed using the product t-norm

{\footnotesize
\begin{equation}
\beta_r(\mathbf{x}) = \prod_{i=1}^{m} A_{ir}(x_i).
\end{equation}
}

\noindent
The consequent function $f_r(\cdot)$ defines the system order, a TS is said to be Zero-order if

\vspace{-0.1cm}
{\footnotesize
\begin{equation}
f_r(\mathbf{x}) = c_r,
\end{equation}
}

\noindent
where $c_r$ is a constant. While a First-order T-S system follows

{\footnotesize
\begin{equation}
f_r(\mathbf{x}) = \sum_{i=1}^{m} w_{ir} x_i + b_r,
\end{equation}
}

\noindent
where $w_{ir}$ are linear coefficients and $b_r$ is the bias term. Then, the overall system output is computed as the weighted average of all rule outputs,

\vspace{-0.2cm}
{\footnotesize
\begin{equation}
y(\mathbf{x}) = \frac{\sum_{r=1}^{R} \beta_r(\mathbf{x}) f_r(\mathbf{x})}{\sum_{r=1}^{R} \beta_r(\mathbf{x})}.
\end{equation}
}
For first-order systems, this becomes

{\footnotesize
\begin{equation}
y(\mathbf{x}) = \frac{\sum_{r=1}^{R} \beta_r(\mathbf{x}) \left(\sum_{i=1}^{m} w_{ir} x_i + b_r\right)}{\sum_{r=1}^{R} \beta_r(\mathbf{x})}.
\end{equation}
}
T-S systems support multiple strategies for parameter estimation,

\noindent
\textbf{Expert Knowledge:} Both antecedent ($\mu_{ir}, \sigma_{ir}$) and consequent parameters ($w_{ir}, b_r$) can be specified based on domain expertise.

\noindent
\textbf{Machine Learning:} Antecedent parameters can be estimated using clustering algorithms (e.g., fuzzy c-means), while consequent parameters are optimized via linear methods such as least squares estimation, owing to the linear-in-parameters form of the consequent functions.  As proposed by Jang \cite{jang1993anfis}, antecedent parameters can also be refined using gradient descent, while consequents are estimated through least squares estimation (LSE). The later architecture, referred to as ANFIS \cite{jang1993anfis} has thousands of extensions in the literature, whether heuristic, derivative-based, or hybrid techniques to optimize hyperparameters, as classified by Karaboga \cite{karaboga2019adaptive}.

\subsection{Explainability Taxonomies}

Preservation of explainability in fuzzy systems implies careful design considerations in the implementation pipeline. Many researchers thoroughly investigated this line of research 
\cite{gacto2010integration, moral2021explainable,
      guillaume2001designing, zhou2008low, gacto2011interpretability}. General direction classifies fuzzy systems explainability into structural and semantic explainability. The former address the model’s complexity—such as the number of rules—while the latter looks at the meaningfulness of fuzzy sets. Table \ref{tab:interpretability} depicts selected key explainability considerations and representative measures used to maintain both types of interpretability.

\begin{table}[t]
\scriptsize
\raggedright
\caption{\small Interpretability Types with Key Considerations and Measures}
\label{tab:interpretability}
\begin{tabular}{p{0.1\textwidth} p{0.13\textwidth} p{0.19\textwidth}}
\toprule
\textbf{Type} & \textbf{Considerations} & \textbf{Measures / Constraints} \\
\midrule
Structural interpretability & Number of rules, conditions, features, fuzzy sets, rule length & Pruning, rule merging, simplification, feature selection, rule base compression \\[3pt]
Semantic interpretability & Distinguishability, coverage, normality, completeness, consistency, linguistic coherence & Parameter tuning, membership function shaping, overlap control, constraint enforcement, merging \\
\bottomrule
\end{tabular}
\end{table}

To maintain structural interpretability in fuzzy systems, limiting the size of the rule base is essential. Semantic interpretability is often maintained by imposing constraints on membership functions. Both types can usually be maintained during design, training, or post-processing. Among the different semantic properties, distinguishability is one of the most common, as it ensures that fuzzy sets are sufficiently distinct and can be assigned to unique linguistic terms \cite{moral2021explainable}. Several approaches quantify distinguishability with the overarching goal of achieving reasonable separation between fuzzy sets. Let $A$ and $B$ denote two fuzzy sets with membership functions $\mu_A(x)$ and $\mu_B(x)$. Jaccard index $J(A,B)=|A \cap B|/|A \cup B|$ evaluates their overlap, with lower values indicating higher distinguishability. In \cite{jin2000fuzzy} and \cite{jin2003generating} Jin et~al. proposed a Gaussian-based measure, where each set is characterized by a center $\mu_i$ and spread $\sigma_i$, computing distinguishability as {\footnotesize$S(A,B)=\sqrt{(\mu_1-\mu_2)^2+(\sigma_1-\sigma_2)^2}$} and defining distinguishability as $d(A,B)=1/(1+S(A,B))$, where a smaller similarity corresponds to greater separability. Mencar~\cite{mencar2007distinguishability} suggested a possibility-based criterion, $P(A,B)=\sup_{x \in U} \min(\mu_A(x),\mu_B(x))$, with values near zero reflecting higher distinguishability.

Several studies have implemented simplification techniques to preserve interpretability. For example, researchers in \cite{guillaume2003new} and \cite{guillaume2004generating} proposed methods for generating interpretable fuzzy rules by removing unnecessary rules, thus reducing the number of membership functions and variables. Thereby preserving structural interpretability and indirectly preserving distinguishability and coverage by merging fuzzy sets based on a sophisticated distance function. Additionally, Gacto et al. \cite{gacto2010integration} developed the GM3M index, a post-processing measure to preserve semantic interpretability by maintaining the original shape of membership functions during tuning, using metrics like displacement, lateral amplitude rate, and area similarity to ensure fuzzy partitions remain interpretable. Other efforts explored graph-simplification \cite{fuchs2020graph} based on merging. These methods have proven to be effective to achieve distinguishability given a set threshold, but fine-tuning a trained model to align with distinguishability may restrict the search space to the trained model rather than the larger parameter space.

Moreover, evolutionary MOO are commonly integrated with fuzzy systems, which can minimize rule base size, impose semantic constraints and maximize performance objectives simultaneously within a MOO
framework \cite{shukla2012review, marquez2010multi, liu2007novel}. In such settings, the conflicting nature of predictive performance and semantic interpretability typically results in a Pareto front of non-dominated solutions. For example, Ishibuchi et al. \cite{ishibuchi2002selecting} used a genetic algorithm for rule selection in classification problems, minimizing the number of rules while maximizing classification accuracy. Pulkkinen and Koivisto proposed a multi-objective genetic fuzzy system\cite{pulkkinen2009dynamically} that ensures distinguishability and coverage while tuning membership functions, learning rules, and adjusting granularities through Wang-Mendel and decision-tree initialization, then applying dynamic $\alpha$-condition (limiting membership intersections) and $\beta$-condition (ensuring strong coverage) constraints during evolution for three-parameter tuning of bell-shaped functions. These methods, while effective in balancing accuracy and explainability, are known to be computationally demanding.

Few researchers have also explored gradient-based MOO for explainable fuzzy systems. In \cite{de1999towards} and \cite{de2002semantic}, semantic constraints were incorporated into MOO through penalty terms. In this $\epsilon$-constrained scheme, the primary objective is augmented with non-equality semantic constraints by adding penalty terms, resulting in a composite objective function

{\footnotesize
\begin{equation}
\bar{J} = J + K_1 J_1 + K_2 J_2,
\end{equation}
}

\noindent
where $J$ is a performance function, such as mean squared error, $J_1$ and $J_2$ represent differentiable functions for coverage and distinguishability. This formulation preserves interpretability by penalizing violations of semantic constraints.

\subsection{The Non-Convex Performance-Explainability Trade-off}

Jointly optimizing performance and explainability in fuzzy systems faces a fundamental challenge: while distinguishability metrics can be convex, the performance objective exhibits non-linearities producing a non-convex multi-objective landscape that weighted scalarization cannot fully capture.

\noindent
Let $\theta = \{c_{ji}, \sigma_{ji}\}$ denote all Gaussian membership parameters for $R$ rules across $n$ input features, where $c_{ji}$ and $\sigma_{ji}$ denote the center and width of the $i$th antecedent of rule $j$. For an input vector $\mathbf{x} = (x_1,\dots,x_n)$, the firing strength of rule $j$ is

\vspace{-0.1cm}
{\footnotesize
\begin{equation}
w_j(\mathbf{x};\theta)
=
\prod_{i=1}^{n}
\exp\!\left(
    -\frac{(x_i - c_{ji})^2}{2\sigma_{ji}^2}
\right),
\end{equation}
}

\noindent
where gradients are expressed as
{\footnotesize
\begin{align}
\frac{\partial \mathcal{L}}{\partial \sigma_{ji}}
&=
\frac{\partial \mathcal{L}}{\partial w_j}\,
w_j\,
\frac{(x_i - c_{ji})^2}{\sigma_{ji}^3}, &
\frac{\partial \mathcal{L}}{\partial c_{ji}}
&=
\frac{\partial \mathcal{L}}{\partial w_j}\,
w_j\,
\frac{(x_i - c_{ji})}{\sigma_{ji}^2}.
\end{align}
}

\noindent
The cubic dependence $\sigma_{ji}^{-3}$ in the width gradient creates orders-of-magnitude scale differences from explainability objectives. In ANFIS, where consequent parameters are eliminated via least squares, normalized firing strengths for each sample $t$ are

\vspace{-0.2cm}
{\footnotesize
\begin{equation}
\bar w_j^{(t)}(\theta)
= \frac{w_j^{(t)}(\theta)}{\sum_{k=1}^{R} w_k^{(t)}(\theta)} ,
\end{equation}
}

\noindent
which form the design matrix $\Phi(\theta)\in\mathbb{R}^{N\times R}$. For a regression problem with mean squared error, the loss becomes

{\footnotesize
\begin{equation}
J(\theta) = \frac{1}{N} \, 
y_{\scriptscriptstyle 1 \times N}^\top 
\Big[ I_{\scriptscriptstyle N \times N} - 
\Phi_{\scriptscriptstyle N \times R} 
\big( \Phi_{\scriptscriptstyle N \times R}^\top \Phi_{\scriptscriptstyle N \times R} \big)^{-1} 
\Phi_{\scriptscriptstyle N \times R}^\top \Big] 
y_{\scriptscriptstyle N \times 1},
\end{equation}
}

\noindent
where the orthogonal projector $\Phi(\theta)\big(\Phi(\theta)^\top\Phi(\theta)\big)^{-1}\Phi(\theta)^\top$ becomes a non-linear rational function of $\theta$ through exponential dependencies, creating multiple local minima despite convex inner least-squares. This non-convexity in $J(\theta)$ prohibits gradient-based weighted scalarization, as such methods can only recover Pareto-optimal solutions when the Pareto front itself is convex~\cite{das1997closer,koski1985defectiveness,miettinen1999nonlinear}—without convexity, they generate only points on the convex hull. Moreover, as membership functions proliferate to preserve distinguishability across finer fuzzy partitions, the required reduction in $\sigma_{ji}$ causes the gradient term $\sigma_{ji}^{-3}$ to explode, drastically amplifying scale discrepancies between performance and explainability objectives.

\section{Methodology}

The proposed explainable Adaptive Neuro-Fuzzy System (X-ANFIS) decouples the performance and explainability objectives rather than combining them into a single aggregate loss function. The alternating bi-objective optimization scheme is summarized in Algorithm~\ref{alg:xanfis} and comprises three sequential phases within each training epoch: (1) \textbf{forward pass} that computes Cauchy membership functions and updates the consequent parameters via regularized least squares estimation, (2) \textbf{backward pass} that updates antecedent parameters by minimizing mean squared error (MSE) through gradient descent, and (3) \textbf{explainability pass} (X-pass) that refines antecedent parameters by minimizing the squared error between pairwise distinguishabilities and a predetermined target distinguishability, based also on gradient descent.

\begin{algorithm}[htbp]
\caption{X-ANFIS: Alternating Optimization}
\label{alg:xanfis}
\fontsize{8}{8}\selectfont
\begin{algorithmic}[1]
\STATE \textbf{Init:} $\mathbf{c}, \mathbf{s} \leftarrow \text{FCM}(\mathbf{X})$
\FOR{$t = 1 \dots T$}
    \STATE \emph{// 1. Forward Pass \& Consequent Optimization}
    \STATE $\mathbf{M} \leftarrow \text{Cauchy}(\mathbf{X}, \mathbf{c}, \mathbf{s})$; \quad $\beta \leftarrow \mathbf{M} / \sum \mathbf{M}$
    \STATE $\mathbf{w} \leftarrow (\beta^T \beta + \lambda \mathbf{I})^{-1} \beta^T \mathbf{y}$ \quad \emph{// Regularized LSE}
    
    \STATE \emph{// 2. Backward Pass (Performance Update)}
    \STATE $\mathcal{L}_{perf} = \text{MSE}(\beta \mathbf{w}, \mathbf{y})$
    \STATE $[\mathbf{c}, \mathbf{s}] \leftarrow [\mathbf{c}, \mathbf{s}] - \eta_{b} \nabla_{\mathbf{c},\mathbf{s}} \mathcal{L}_{perf}$ \quad \emph{// Standard Gradient Descent}
    
    \STATE \emph{// 3. X-Pass (Explainability Update)}
    \STATE $\mathcal{J}_{pair} = \{(i,j) \mid \text{rank}(c_{i,f}) + 1 = \text{rank}(c_{j,f})\}$ \quad \emph{// Adjacent sets}
    \STATE $\mathbf{c} \leftarrow \mathbf{c} - \eta_{x} \sum_{f} \sum_{(i,j) \in \mathcal{J}} \nabla_{\mathbf{c}} |D(c_i, c_j) - D_{target}|$
    \STATE \emph{// 4. Validation \& Early Stopping}
    \STATE \textbf{break if val MSE has not improved for $p$ epochs}
\ENDFOR
\RETURN $\mathbf{c}, \mathbf{s}, \mathbf{w}$
\end{algorithmic}
\end{algorithm}

\subsection{Cauchy Forward Pass}
The structural interpretability of ANFIS can be controlled to some extent prior to training by considering a minimum number of features and clusters, but the semantic interpretability remains vulnerable to fundamental initialization-based bottlenecks. The spread term $\sigma_{ji}^{-3}$ from equation (10) can imply two stability challenges in the gradient expression:

\subsubsection{Exploding Gradients for Small Initialization} when setting spread to small values to initiate high distinguishability, the $\sigma_{ji}^{-3}$ term dominates the gradient

{\footnotesize
\begin{equation}
\left|\frac{\partial \mathcal{L}}{\partial \sigma_{ji}}\right| 
\propto \frac{1}{\sigma_{ji}^3} \to \infty \quad \text{as } \sigma_{ji} \to 0.
\end{equation}}

\noindent
This leads to unstable parameter updates and requires aggressive gradient clipping or extremely low learning rates.

\subsubsection{Vanishing Updates for Large Initialization}
Conversely, when spread parameters are initialized with large values, $|\partial \mathcal{L}/\partial \sigma_{ji}| \propto \sigma_{ji}^{-3} \to 0$ as $\sigma_{ji} \to \infty$, causing negligible spread updates and lower distinguishability.
Consequently, converging to separable membership functions with gradient descent could be highly unstable in classical Gaussian-based ANFIS models, whereas many performance-oriented research efforts opt to initiate spread to larger values for performance and stability gains \cite{cui2021curse, cui2022pytsk, wu2019optimize}. Therefore, we adopt the Cauchy membership function in neuro-fuzzy system design

{\footnotesize
\begin{equation}
\mu(x) \;=\; \frac{1}{1 + \left(\dfrac{x-c}{\gamma}\right)^2},
\end{equation}
}
\noindent
where \(c\) denotes the center and \(\gamma\) the scale of the function. Unlike the Gaussian function, which exhibits exponential decay, the Cauchy function decays algebraically, resulting in heavier tails that can be more stable when scale parameters are initialized to small values.

To derive the gradients for backpropagation, let \(\mathcal{L}\) denote the scalar loss and \(\partial \mathcal{L}/\partial \mu(x)\) represent the upstream gradient. Applying the chain rule:

{\footnotesize
\begin{equation}
\frac{\partial \mu}{\partial \gamma}
= -\!\left(1+\frac{(x-c)^2}{\gamma^2}\right)^{-2}\!\cdot\!\left(-\frac{2(x-c)^2}{\gamma^3}\right)
= 2\,\mu^{2}(x)\,\frac{(x-c)^{2}}{\gamma^{3}},
\end{equation}
}
\noindent
noting that $\mu^{2}(x) = \bigl(1 + (x-c)^{2}/\gamma^{2}\bigr)^{-2}$. Thus the partial derivatives of the loss are

{\footnotesize
\begin{equation}
\frac{\partial \mathcal{L}}{\partial \gamma}
= \frac{\partial \mathcal{L}}{\partial \mu}\cdot 2\,\mu^2(x)\,\frac{(x-c)^2}{\gamma^3}, \quad
\frac{\partial \mathcal{L}}{\partial c}
= \frac{\partial \mathcal{L}}{\partial \mu}\cdot 2\,\mu^2(x)\,\frac{(x-c)}{\gamma^2}.
\label{eq:L_gamma}
\end{equation}
}
\vspace{1mm}

\noindent
Having derived the necessary gradients, the first two training steps for X-ANFIS can now be described. In the forward pass, Cauchy membership functions are computed using equation (14), followed by calculation of the firing strength per rule via the product t-norm and its normalization according to equation (11). The consequent parameters of the zero-order TSK model are then estimated using regularized least squares. In the backward pass, the antecedent parameters (centers and scales) are updated using gradient descent with the learning rate \(\eta\):

{\footnotesize
\begin{equation}
\gamma^{(t+1)} = \gamma^{(t)} - \eta \cdot \frac{\partial \mathcal{L}}{\partial \gamma}, \quad
c^{(t+1)} = c^{(t)} - \eta \cdot \frac{\partial \mathcal{L}}{\partial c},
\end{equation}
}
\noindent
where the superscript \((t)\) denotes the iteration number. Although the Cauchy gradient retains the cubic dependency \(\gamma^{-3}\) similar to the Gaussian \(\sigma^{-3}\) term, the algebraic decay of the Cauchy function combined with the moderating effect of the \(\mu^2(x)\) term prevents the gradient explosion when spreads are initialized to low values, thereby enabling stable convergence toward distinguishable membership functions.

\subsection{Explainability Pass}

Explainability pass (X-pass) maintains optimal cluster separation between adjacent fuzzy sets by optimizing a differentiable  objective function. The distinguishability between consecutive sets $i$ and $j$ is defined as:

{\footnotesize
\begin{equation}
D_{ij}^{(f)} = \sqrt{(\mu_i^{(f)} - \mu_j^{(f)})^2 + (\sigma_i^{(f)} - \sigma_j^{(f)})^2},
\end{equation}
}
\noindent
where $\mu_i^{(f)}$ and $\sigma_i^{(f)}$ are the center and scale parameters of the fuzzy set $i$ for the feature $f$, respectively. The indices $i$ and $j$ denote adjacent fuzzy sets in the partition. Since lower distinguishability implies larger overlap between fuzzy sets, larger values are recommended to maintain reasonable overlap, while excessively large values could diminish overlap and violate the coverage condition. The appropriate distinguishability threshold ultimately depends on the desired degree of explainability; in practice, researchers have adopted thresholds greater than approximately 0.4 \cite{jin2000fuzzy, khaled2025interpretable}. In this study, we adopt a target distinguishability range of [0.4–0.5] to balance semantic clarity and adequate coverage. To regulate this balance, the X-pass optimizes pairwise distinguishability by minimizing a squared-error objective, which provides a smooth and differentiable penalty that increases more strongly when adjacent sets drift too close or too far apart. The corresponding gradient updates for adjacent sets are

\vspace{-4mm}
{\footnotesize
\begin{equation}
L_{\text{X-pass}} = \tfrac{1}{2}(D_{ij}^{(f)} - D_{\text{target}})^2
\end{equation}
\vspace{-2mm}
\begin{align}
\frac{\partial L}{\partial \mu_i} &= \frac{D - D_{\text{target}}}{D} (\mu_i - \mu_j), \quad
\frac{\partial L}{\partial \mu_j} = \frac{D - D_{\text{target}}}{D} (\mu_j - \mu_i), \notag\\
\frac{\partial L}{\partial \sigma_i} &= \frac{D - D_{\text{target}}}{D} (\sigma_i - \sigma_j), \quad
\frac{\partial L}{\partial \sigma_j} = \frac{D - D_{\text{target}}}{D} (\sigma_j - \sigma_i).
\end{align}
}

Using these gradients, the X-pass update is applied exclusively to the centers of the cluster $\mu$, while the scale parameter $\sigma$ remains frozen. This constraint prevents trivial minimization of widths that would violate coverage. Consequently, in the general flow of X-ANFIS, antecedent parameters ($\mu$, $\sigma$) are first initialized via Fuzzy C-Means (FCM) clustering, the backward pass then updates the parameters towards the error minimum, potentially leading to larger overlap; subsequently, the X-pass applies a corrective pairwise adjustment to the centers, restoring the locally desired separation dictated by $D_{\text{target}}$. The parameter update protocol in all training phases is detailed in Table \ref{tab:xpass_optimization}.

\begin{table}[htbp]
\scriptsize 
\centering
\caption{\small Parameter Update Protocol for Alternating Bi-objective Optimization of X-ANFIS}
\label{tab:xpass_optimization}
\renewcommand{\arraystretch}{1.2}
\begin{tabular}{|p{1.2cm}|p{1cm}|p{1.2cm}|p{0.9cm}|p{1cm}|p{0.7cm}|}
\hline
\textbf{} & \textbf{Parameter} & \textbf{Initialization} & \textbf{Forward Pass} & \textbf{Backward Pass} & \textbf{X-Pass} \\
\hline
\textbf{Antecedents} & $\mu$ & FCM & Fixed$^{**}$ & GD & GD \\
\cline{2-6}
& $\sigma$ & FCM & Fixed & GD & Fixed \\
\hline
\textbf{Consequents} & $w$ & N/A* & LSE+$L_2$ & Fixed & Fixed \\
\hline
\end{tabular}
\\[2mm]
\parbox{\linewidth}{\scriptsize 
* Consequents are obtained via closed-form least-squares; no initialization required.\\
** Fixed indicates that the parameter is not updated.
}
\end{table}

\noindent
This alternating structure isolates the competing objectives: backward pass optimizes predictive accuracy without explainability constraints, while X-pass enforces semantic interpretability by reshaping the partition where fuzzy sets become indistinguishable.  Convergence is determined by validation MSE through early stopping with patience $p$ epochs, ensuring that the model terminates based on generalization performance rather than distinguishability convergence.

\section{Evaluation}
X-ANFIS is evaluated through three experimental settings, initialization experiments, Pareto experiments and heatmap experiments as summarized in Table \ref{tab:exp_hyper_manual}. Training includes a total of nine regression datasets from the UCI Machine Learning Repository, spanning 308 to 35,040 samples and 4 to 81 features across domains including energy systems, material science, and transportation as shown in Table~\ref{tab:datasets}. Firstly, initialization experiments examine the stability of Gaussian and Cauchy membership functions across six spread values. Using 30 rules on the Combined Cycle Power Plant dataset, the analysis tracks how the training trajectory evolves as spreads are initialized to small values that promote high distinguishability. Secondly, Pareto experiments compare three modeling approaches across all datasets: classical single-objective ANFIS with Gaussian MFs, gradient-based scalarized MO ANFIS with Cauchy MFs over 500 different weights assigned to distinguishability in the range of 0.01 to 10 and the proposed X-ANFIS with Cauchy MFs. In these experiments, structural interpretability is preserved solely by the rule count, with each model limited to five rules, while semantic interpretability addresses distinguishability, coverage, consistency and completeness. Other structural characteristics—such as the number of features and rule length, are intentionally excluded from the scope of this study, as they are typically managed through simplification techniques. Thirdly, heatmap experiments visualize MF distributions using 10 rules on the combined power plant dataset to illustrate the spatial organization of fuzzy partitions produced by each method, with emphasis on consistency. Shared hyperparameters consist of learning rate of 0.1, L2 regularization ($\lambda = 10^{-4}$), min–max scaling to $[0,1]$, gradient clipping $[-1,1]$, and a 70/10/20 data split for training, validation, and testing. Completeness and coverage—semantic properties ensuring every input has non-zero membership and fuzzy sets collectively span the input space—are promoted by bounding centers and scale to the normalized input range $[0,1]$. Performance is evaluated using the coefficient of determination ($R^2$), Mean Absolute Error (MAE), Mean Squared Error (MSE), Root Mean Squared Error (RMSE), and mean distinguishability ($D$) between adjacent fuzzy sets.

\begin{table}[htbp]
\centering
\caption{\small Summary of Experimental Design}
\label{tab:exp_hyper_manual}
\renewcommand{\arraystretch}{1.1}
\scriptsize

\begin{tabular}{|c|p{1.5cm}|p{1.2cm}|c|p{1.2cm}|}
\hline
\textbf{No} & \textbf{Model} & \textbf{MF Type} & \textbf{R} & \textbf{Experiments} \\
\hline

\multicolumn{5}{|l|}{\textbf{Initialisation Experiments}} \\
\hline
1 & ANFIS & Gaussian & 30 & 6 \\
2 & ANFIS & Cauchy & 30 & 6 \\
\hline

\multicolumn{5}{|l|}{\textbf{Pareto Experiments}} \\
\hline
3 & ANFIS & Gaussian & 5 & 9 \\
4 & MO-ANFIS & Cauchy & 5 & 9x500 \\
5 & X-ANFIS & Cauchy & 5 & 9 \\
\hline

\multicolumn{5}{|l|}{\textbf{Heatmap Experiments}} \\
\hline
6 & ANFIS & Gaussian & 10 & 1 \\
7 & X-ANFIS & Cauchy & 10 & 1 \\
8 & MO-ANFIS & Cauchy & 10 & 1x500 \\
\hline

\end{tabular}
\end{table}

\begin{table}[b]
\scriptsize
\centering
\renewcommand{\arraystretch}{1.3} 
\setlength{\tabcolsep}{4pt}       
\caption{\small Summary of Regression Real-world Datasets Used in Experiments Obtained from UCI Machine Learning Repository}
\begin{tabular}{|c|l|c|c|l|}
\hline
\textbf{\#} & \textbf{Dataset Name} & \textbf{Samples} & \textbf{Features} & \textbf{Prediction Target} \\ \hline
1  & Airfoil Self-Noise            & 1,503   & 5  &  Sound pressure \\ \hline
2  & Concrete Compressive Strength & 1,030   & 8  &  Concrete strength \\ \hline
3  & Energy Efficiency             & 768     & 8  & Heating and cooling  \\ \hline
4  & Yacht Hydrodynamics           & 308     & 6  &  Sailing resistance \\ \hline
5  & Bike Sharing (Day)            & 731     & 13 &  Daily count \\ \hline
6  & Bike Sharing (Hour)           & 17,379  & 16 &  Hourly count \\ \hline
7  & Steel Industry Energy         & 35,040  & 9  &  Energy consumption \\ \hline
8  & Combined Cycle Power Plant    & 9,568   & 4  &  Energy output \\ \hline
9  & Superconductivity             & 21,263  & 81 & Critical temperature  \\ \hline
\end{tabular}
\label{tab:datasets}
\end{table}

\subsection{Parameter Update Trajectory Analysis}

\begin{figure*}[htbp]  
    \centering
    \includegraphics[width=1\textwidth]{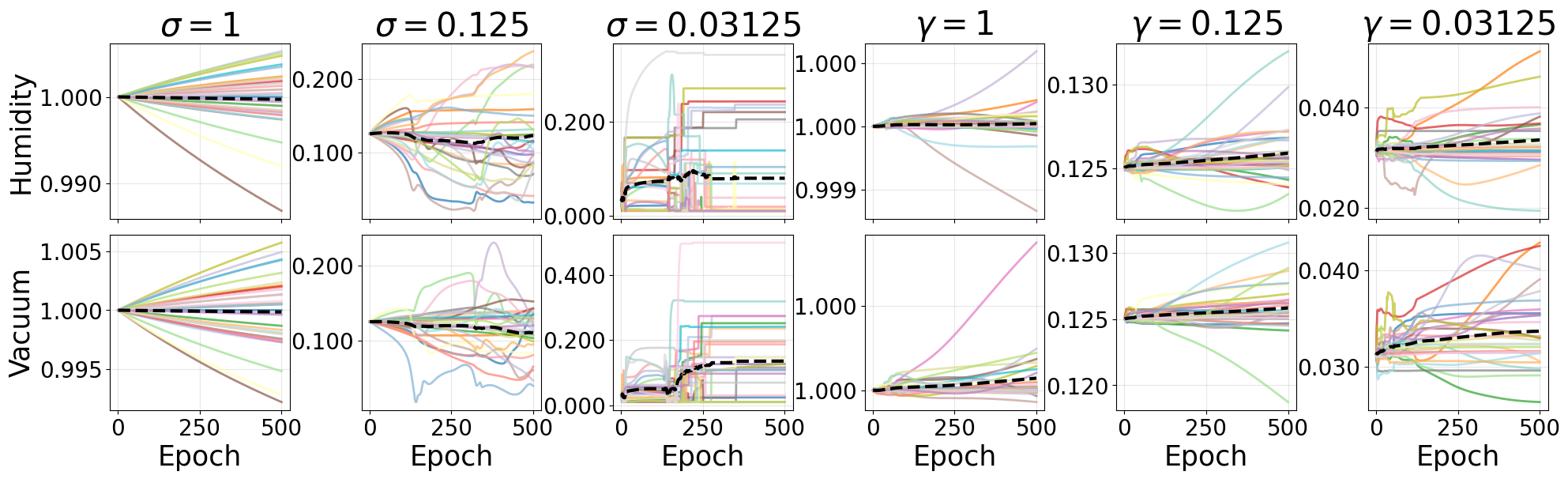}  
    \caption{\small Effect of MF initialization on ANFIS parameter update trajectories for the Combined Cycle Power Plant dataset, for humidity and vacuum features. Columns correspond to different initialized spread values $\sigma$/$\gamma$ values; left three columns are Gaussian MFs, right three are Cauchy MFs.}
    \label{fig:init}
\end{figure*}

To examine initialization stability, we trained six ANFIS models with Gaussian MFs and six with Cauchy MFs, using spread values that decrease exponentially from $1$ to $0.03125$ under fixed hyperparameters. Table \ref{tab:anfis_init_performance} presents the performance of all models, while Figure \ref{fig:init} illustrates representative spread update trajectories for two input features, displaying three cases each.
Gaussian-based models exhibited strong sensitivity to initialization: when $\sigma \le 0.125$, the parameter trajectories tend to be non-monotonic, as depicted in their plot, and non-convergent below $0.125$, as shown in the performance results. This instability is reflected in test metrics, where the MSE rose from about $0.003$ to $0.25$ and $R^{2}$ dropped to nearly $-4$ for $\sigma \le 0.0625$. In contrast, Cauchy-based models maintained relatively smooth trajectories across all initialization scales, with trajectory behavior and test errors largely unchanged. Trajectories remained almost monotonic even at the lowest $\gamma$ value, as depicted in the mean trajectory.  Specifically, this robustness arises from the heavy-tailed Cauchy activation, which maintains non-zero rule firing even for narrow spreads. Since narrow spreads enhance semantic interpretability by improving MF distinguishability, the stability of the Cauchy model under small-scale further builds the case for its suitability in interpretable Neuro-Fuzzy modeling.

\begin{table}[htbp]
\renewcommand{\arraystretch}{1.2} 
\centering
\scriptsize
\caption{\small Test performance of Gaussian and Cauchy ANFIS models under different initialization scales}
\label{tab:anfis_init_performance}
\begin{tabular}{c c c c c c}
\hline
\textbf{MF Type} & \textbf{Initialized Std/Scale} & \textbf{MSE} & \textbf{RMSE} & \textbf{MAE} & \textbf{R²} \\
\hline
\multirow{6}{*}{Gaussian}
& 1      & 0.0032 & 0.0563 & 0.0446 & 0.9377 \\
& 0.5    & 0.0030 & 0.0546 & 0.0428 & \textbf{0.9413} \\
& 0.25   & 0.0029 & 0.0539 & 0.0421 & \textbf{0.9428} \\
& 0.125  & 0.0038 & 0.0614 & 0.0487 & 0.9260 \\
& 0.0625 & 0.2513 & 0.5013 & 0.4477 & -3.9394 \\
& 0.03125& 0.2490 & 0.4990 & 0.4448 & -3.8937 \\
\hline
\multirow{6}{*}{Cauchy}
& 1      & 0.0031 & 0.0556 & 0.0438 & \textbf{0.9393} \\
& 0.5    & 0.0032 & 0.0563 & 0.0447 & 0.9377 \\
& 0.25   & 0.0031 & 0.0561 & 0.0442 & 0.9382 \\
& 0.125  & 0.0032 & 0.0567 & 0.0449 & \textbf{0.9369} \\
& 0.0625 & 0.0031 & 0.0559 & 0.0443 & \textbf{0.9385} \\
& 0.03125& 0.0031 & 0.0556 & 0.0438 & \textbf{0.9393} \\
\hline
\end{tabular}
\end{table}

\subsection{Performance and Distinguishability}
\begin{figure}[htbp]
    \centering
    \includegraphics[width=1.0\columnwidth]{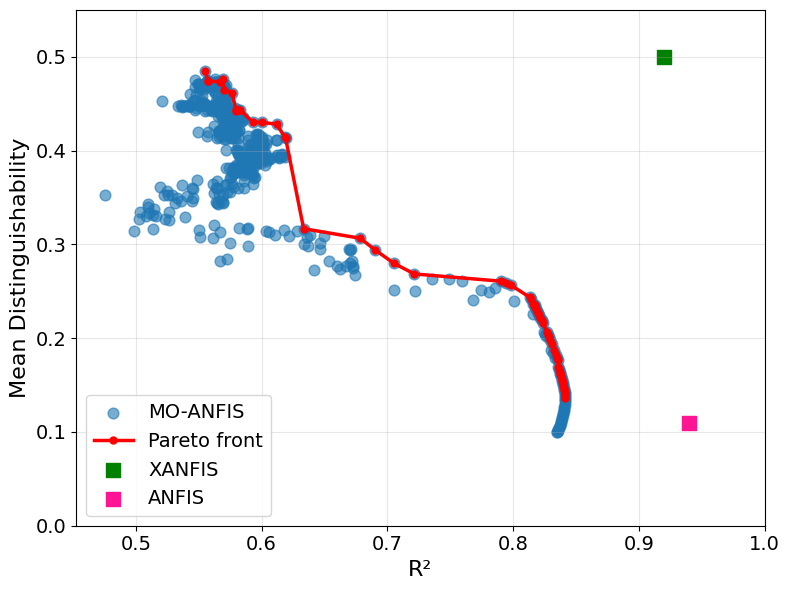}
    \caption{\small Training results of 500 models depicted as R² versus mean distinguishability for ANFIS, X-ANFIS and MO-ANFIS models. MO-ANFIS Pareto front shown in red. Classic ANFIS shown in pink. X-ANFIS shown in green.}
    \label{fig:dist}
\end{figure}

Having established stable training dynamics, we now evaluate whether the alternating bi-objective optimization strategy enables X-ANFIS to achieve superior accuracy-explainability trade-offs compared to baseline approaches. Table \ref{tab: anfis_comparison} summarizes the $R^2$ performance and mean distinguishability ($D$) for all Pareto experiments. In most datasets, conventional ANFIS achieves the highest predictive performance, but its distinguishability remains minimal ($<0.15$). In contrast, while MO-ANFIS increases distinguishability, it does so only at a substantial cost to predictive accuracy. Figure~\ref{fig:dist} visualizes this trade-off, showing the Pareto front -marked in red- generated from 500 MO-ANFIS models alongside ANFIS and X-ANFIS for the combined cycle power plant dataset. ANFIS achieves $R^2 = 0.94$ with $D \approx 0.11$, while MO-ANFIS is able to produce solutions with $D$ greater than 0.25 (half of the target distinguishability) only when $R^2$ drops below 0.8. Overall, X-ANFIS achieves a far more favorable balance between accuracy and explainability, frequently delivering slightly lower performance than ANFIS while attaining the highest distinguishability $D \approx 0.50$ in 8 out of 9 datasets.

\begin{table*}[htbp]
\centering
\setlength{\abovecaptionskip}{2pt}
\setlength{\belowcaptionskip}{1pt}
\caption{\small Performance Comparison of ANFIS, MO-ANFIS and X-ANFIS Across 9 Datasets, Evaluated Using ($R^2$) and Distinguishability ($D$). 
95\% Confidence Intervals (CI) are Reported for $R^2$ and $D$.}
\label{tab: anfis_comparison}
\vspace{0.3em}
\renewcommand{\arraystretch}{1.2}
\setlength{\tabcolsep}{3pt}
\scriptsize
\rowcolors{3}{gray!3}{white}

\begin{tabular}{
    p{1.8cm}
    *{4}{>{\centering\arraybackslash}p{1.2cm}}
    *{2}{>{\centering\arraybackslash}p{1.5cm}}
    *{4}{>{\centering\arraybackslash}p{1.2cm}}
}
\toprule
\multirow{2}{*}{\textbf{Dataset}} 
& \multicolumn{4}{c}{\textbf{ANFIS}} 
& \multicolumn{2}{c}{\textbf{MO-ANFIS}} 
& \multicolumn{4}{c}{\textbf{X-ANFIS}} \\
\cmidrule(lr){2-5} \cmidrule(lr){6-7} \cmidrule(lr){8-11}
& $\mathbf{R^2}$ & $\mathbf{D}$ & $\mathbf{CI\ R^2}$ & $\mathbf{CI\ D}$
& $\mathbf{R^2}$ & $\mathbf{D}$
& $\mathbf{R^2}$ & $\mathbf{D}$ & $\mathbf{CI\ R^2}$ & $\mathbf{CI\ D}$ \\
\midrule

Airfoil Noise
& \textbf{0.56} & 0.15 & 0.53 -- 0.58 & 0.14 -- 0.16
& 0.00 -- 0.18 & 0.12 -- 0.47
& 0.43 & \textbf{0.50} & 0.33 -- 0.51 & 0.49 -- 0.50 \\

Concrete Strength
& \textbf{0.75} & 0.12 & 0.73 -- 0.77 & 0.11 -- 0.13
& 0.00 -- 0.38 & 0.09 -- 0.44
& 0.73 & \textbf{0.50} & 0.66 -- 0.79 & 0.49 -- 0.50 \\

Energy Efficiency
& 0.91 & 0.16 & 0.89 -- 0.93 & 0.15 -- 0.17
& 0.10 -- 0.75 & 0.11 -- 0.46
& \textbf{0.91} & \textbf{0.50} & 0.87 -- 0.93 & 0.49 -- 0.50 \\

Yacht
& \textbf{0.87} & 0.15 & 0.84 -- 0.91 & 0.14 -- 0.18
& -0.30 -- 0.15 & 0.03 -- 0.50
& 0.86 & \textbf{0.50} & 0.78 -- 0.92 & 0.49 -- 0.50 \\

Bike (Day)
& 0.95 & 0.12 & 0.92 -- 0.98 & 0.11 -- 0.14
& -0.30 -- 0.70 & 0.05 -- 0.39
& \textbf{0.98} & \textbf{0.50} & 0.97 -- 0.99 & 0.49 -- 0.50 \\

Bike (Hour)
& \textbf{0.96} & 0.10 & 0.93 -- 0.98 & 0.08 -- 0.11
& -0.16 -- 0.53 & 0.04 -- 0.35
& 0.94 & \textbf{0.50} & 0.93 -- 0.94 & 0.49 -- 0.50 \\

Power Plant
& \textbf{0.94} & 0.11 & 0.93 -- 0.94 & 0.10 -- 0.11
& 0.44 -- 0.84 & 0.09 -- 0.49
& 0.92 & \textbf{0.50} & 0.91 -- 0.93 & 0.49 -- 0.50 \\

Superconductivity
& \textbf{0.62} & 0.08 & 0.56 -- 0.63 & 0.07 -- 0.08
& 0.45 -- 0.52 & 0.06 -- 0.17
& 0.61 & \textbf{0.50} & 0.58 -- 0.62 & 0.49 -- 0.50 \\

Steel Industry
& \textbf{0.35} & 0.16 & 0.34 -- 0.35 & 0.15 -- 0.16
& 0.20 -- 0.34 & 0.16 -- \textbf{0.51}
& 0.34 & 0.50 & 0.32 -- 0.36 & 0.49 -- 0.50 \\

\bottomrule
\end{tabular}
\vspace{-3mm}
\end{table*}

\noindent
\subsection{Heatmap Experiments}
\begin{figure}[htbp]
    \centering
    \includegraphics[width=0.7\columnwidth]{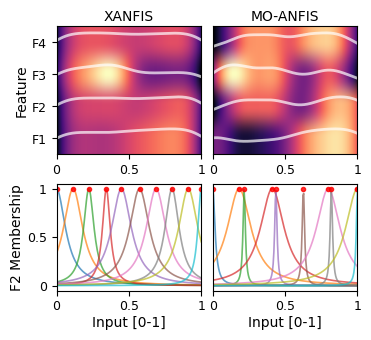}
    \caption{\small Comparison of X-ANFIS and MO-ANFIS. Top row presents Kernel Density Estimation heatmaps of MF centers for features temperature, pressure, humidity, vacuum, named as F1–F4. Bottom row displays the corresponding MF shapes for F2, with red circles indicating center locations.}
    \vspace{-15pt}
    \label{fig:heat}
\end{figure}

To investigate the spatial distribution of the resulting partitions, we re-trained X-ANFIS and MO-ANFIS on the Combined Cycle Power Plant dataset using 10 fuzzy rules, selecting a MO-ANFIS model from the Pareto front with quantitative distinguishability ($D$) matched to X-ANFIS regardless of performance. Figure \ref{fig:heat} contrasts the topological organization of these partitions: the kernel density estimate (KDE) heatmaps (top row) reveal that X-ANFIS maintains a structured, spatially coherent distribution of centers across features F1–F4, manifested as distinct laminar density contours. This regularity translates into the semantic plane (bottom row), where X-ANFIS produces almost equidistant, uniformly distributed Cauchy membership functions satisfying distinguishability, consistency, coverage, and completeness. Conversely, the MO-ANFIS heatmap displays irregular high-density pockets, while the corresponding membership functions suffer from severe spatial clustering affected by data density (e.g., adjacent centers collapsing near inputs 0.2 and 0.4 for F1) and significant coverage gaps. These irregularities also violate the rule consistency condition fundamental to semantic interpretability. The spatial coherence stems from the local optimization structure of X-ANFIS: each adjacent pair of fuzzy sets $(i,j)$ receives an independent gradient update without aggregation between pairs, enforcing distinguishability constraints. In contrast, MO-ANFIS optimizes the global scalarized objective $\mathcal{L} = \mathcal{L}_{\text{perf}} + \alpha \sum_{f,i,j} (D_{ij} - D_{\text{target}})^2$, aggregating all pairwise terms and neglecting the local structure. Although both achieve comparable distinguishability ($D \approx 0.50$), only X-ANFIS satisfies the rule consistency criterion—uniform spacing without clustering or coverage gaps.

\section{Discussion and Conclusion}
Interpretability in fuzzy systems is rarely a single-axis problem. While Pareto experiments focus on a fixed rule count of five, experimenting with larger rule counts revealed a deeper coupling between structural and semantic concerns: attempting to enforce distinguishability across 50 rules within a bounded universe of discourse forces an uncomfortable choice between violating coverage and sacrificing semantic coherence — a tension that grows with model complexity, not input dimensionality. The superconductivity results underscore this point: 81 features did not pose an interpretability challenge, but aggressive rule counts did. Explainable fuzzy models must therefore be designed with both dimensions in mind from the outset, rather than treating them as independent concerns.

However, this complexity does not need to drive the field toward expensive optimization strategies. The assumption that non-convex accuracy-explainability trade-offs necessitate evolutionary methods deserves scrutiny. This work shows that gradient-based methods, when decoupled rather than scalarized, can reach regions of the Pareto front that weighted approaches fundamentally cannot. Beyond this, the stability findings around Cauchy membership functions reveal a quieter insight: the field's preference for large-spread initializations has been trading away semantic interpretability before optimization even begins. Explainability need not be a post-hoc correction; it can be woven into the gradient flow itself.  Future work will extend this framework to Mamdani-type systems to examine whether these principles generalize beyond the Takagi-Sugeno paradigm.

\section*{Acknowledgment}
\vspace{-0.5mm}
{\footnotesize
This publication is part of the Innovation Lab for Utilities on Sustainable Technology and Renewable Energy project (ILUSTRE), which is partially funded by the Dutch Research Council (NWO).
}
\vspace{-5mm}

\section*{}
\balance
\bibliographystyle{IEEEtran}
\bibliography{ref}

\end{document}